\documentclass[10pt,twocolumn,letterpaper]{article}

\usepackage{cvpr}
\usepackage{makecell}

\usepackage{color}

\usepackage[dvipsnames]{xcolor}

\definecolor{julieta_colour}{RGB}{217,95,2} %

\definecolor{cvprblue}{rgb}{0.21,0.49,0.74}
\usepackage[pagebackref,breaklinks,colorlinks,citecolor=cvprblue]{hyperref}

\def\paperID{\#265} 
\def\confName{3DV\xspace}
\def\confYear{2026\xspace}

\title{TurboPortrait3D: Single-step diffusion-based \\fast portrait novel-view synthesis}

\author{Emily Kim\\
Carnegie Mellon University\\
{\tt\small ekim2@andrew.cmu.edu}\\
\and
Julieta Martinez\\
Meta Reality Labs\\
{\tt\small julietamartinez@meta.com}
\and
Timur  Bagautdinov\\
Meta Reality Labs\\
{\tt\small timurb@meta.com}
\and
Jessica Hodgins\\
Carnegie Mellon University\\
{\tt\small jkh@andrew.cmu.edu}
}

\begin{document}
\maketitle
\begin{abstract}
We introduce TurboPortrait3D: a method for low-latency
novel-view synthesis of human portraits.
Our approach builds on the observation that existing image-to-3D 
models for portrait generation, while capable of producing renderable 
3D representations, are prone to visual artifacts, 
often lack of detail, and tend to fail at fully preserving the identity of the subject.
On the other hand, image diffusion models excel at generating
high-quality images, but besides being computationally expensive, are not grounded in 3D and thus
are not directly capable of producing multi-view consistent outputs.

In this work, we demonstrate that image-space diffusion models
can be used to significantly enhance the quality of existing
image-to-avatar methods, while maintaining 3D-awareness and running with low-latency.
Our method takes a single frontal image of a subject as input, and applies a feedforward
image-to-avatar generation pipeline to obtain an initial 3D representation and
corresponding noisy renders.
These noisy renders are then fed to a single-step diffusion model
which is conditioned on input image(s), and is specifically trained to refine 
the renders in a multi-view consistent way.
Moreover, we introduce a novel effective training strategy that includes pre-training
on a large corpus of synthetic multi-view data, followed by fine-tuning on 
high-quality real images.
We demonstrate that our approach both qualitatively 
and quantitatively outperforms current state-of-the-art
for portrait novel-view synthesis, while being efficient in time.
\end{abstract}    
\section{Introduction}

\begin{figure}
    \centering
    \includegraphics[width=0.8\linewidth]{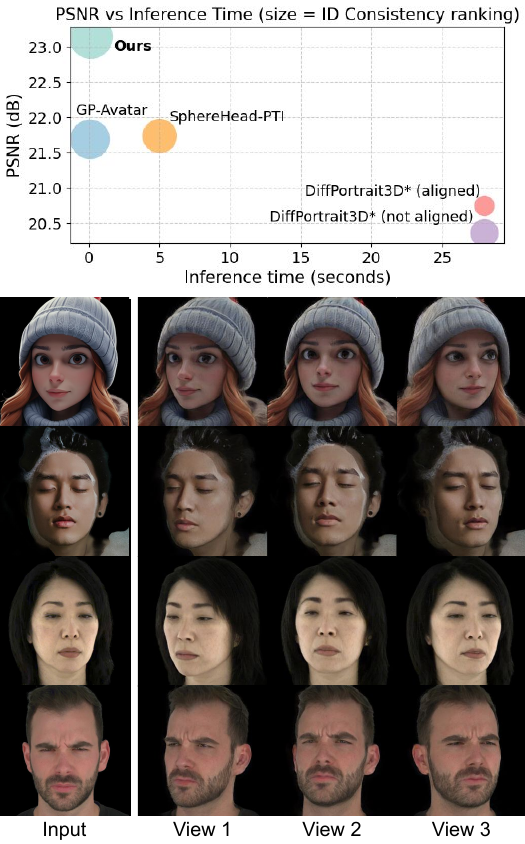}
     \caption{
  \textbf{Teaser -- }
  From a single frontal reference image, TurboPortriait3D is able to generate 3D-aware novel portrait views. Unoptimized, our approach does this at about 120 ms, or roughly 8 frames per second.
  }
    \label{fig:placeholder}
\end{figure}

Fast, identity-faithful portrait novel view synthesis is essential for telepresence, live content creation, and animation workflows, where multi-view consistency and perceptual quality must be maintained under strict latency constraints. While novel view synthesis has been extensively studied, producing high-quality, 3D-aware portraits from a single input image remains challenging. Human faces have structured yet diverse geometry, significant identity variation, and highly expressive motion~\cite{cai2023faces}, and human perception is particularly sensitive to visual artifacts in this domain, making this a high-impact but demanding benchmark for generative modeling.

Existing 3D-aware novel view synthesis methods, including both diffusion- and GAN-based approaches, have complementary strengths and limitations. Diffusion-based methods~\cite{gu2024diffportrait3d} can achieve high visual quality but require many iterative denoising steps, making them unsuitable for real-time use. GAN-based methods avoid iterative refinement and run faster, but often suffer from geometric distortions and require computationally expensive latent inversion~\cite{roich2021pivotal}. Moreover, both typically operate in 2D or implicit 3D space, generating each novel view independently and risking cross-view inconsistencies.

Avatar generation models offer a more efficient alternative by directly predicting explicit, parametric 3D avatars from a single image via feed-forward encoders~\cite{chu2024gpavatar}. These models enable re-rendering from arbitrary viewpoints and animating with consistent geometry without re-running the generator. However, they often lose fine-grained identity cues and struggle under extreme poses or expressions, limiting visual fidelity.

We introduce TurboPortrait3D, a feed-forward, single-step diffusion framework for refining 3D-consistent novel views from feed-forward avatar generators (\eg, GP-Avatar). Given a small set of low-quality but 3D-consistent renders, our method produces high-fidelity, identity-preserving results in a single learned denoising step. Unlike generic single-step refinement methods (e.g., Difix3D~\cite{wu2025difix3d}), TurboPortrait3D is designed specifically for the portrait domain, where subtle identity features, hair details, and expression transfer are critical. Our framework integrates an attention reshaping block for multi-view fusion, a variable-noise training regime with fixed-noise inference for stability, and a memory-efficient batch design that enables low-latency synthesis (\eg, two views in 120 ms at $256\times256$) while maintaining 3D-aware consistency.

We evaluate TurboPortrait3D on Ava-256 and NeRSemble using non-overlapping identity splits, training on a mix of synthetic and real portrait data spanning diverse poses and expressions. Our contributions are:
\begin{enumerate}
\item The first single-step diffusion framework for refining 3D-consistent novel view portraits from avatar synthesis models, significantly improving visual fidelity and identity preservation.
\item A variable-noise training strategy for robustness to varying input quality, with fixed-noise during inference for stable outputs.
\item A memory-efficient multi-view refinement architecture that supports low-latency synthesis while preserving 3D-awareness across viewpoints. 
\end{enumerate}
\section{Related Work}
\label{sec:related}

We focus on methods for generating 3D-aware synthetic portraits, which can be broadly categorized into optimization-based approaches and avatar generation-based approaches.

\paragraph{Optimization-based novel view synthesis.}
Several works have tackled the problem of synthesizing 3D-aware portraits from a single 2D image. Among GAN-based methods, EG3D~\cite{Chan2021} generates 3D-aware faces from in-the-wild images using a tri-plane latent representation and an efficient neural renderer. PanoHead~\cite{an2023panohead} extends this framework to full 360\textdegree~head generation by incorporating a small set of hairstyle captures~\cite{kim2021k} and increasing the number of feature planes to improve expressivity. SphereHead~\cite{li2024spherehead} further enhances realism by introducing a spherical coordinate system into the tri-plane framework, reducing geometric artifacts. While these methods can generate novel views from a single reference image, they rely on computationally intensive optimization processes to invert the input into the GAN’s latent space.

Diffusion-based methods offer an alternative approach to novel view synthesis. A common strategy involves using ControlNet~\cite{zhang2023adding} to guide the generation of viewpoint and head pose. To preserve subject identity, various conditioning mechanisms have been proposed. Arc2Face~\cite{paraperas2024arc2face}, for example, encodes identity features using ArcFace~\cite{deng2019arcface} and embeds them as CLIP-style condition vectors. DiffPortrait3D~\cite{gu2024diffportrait3d} introduces an auxiliary UNet to encode identity features, which are fused with the main diffusion process via cross-attention. While these methods generate high-quality outputs, they typically require 25 to 50 iterative denoising steps during inference, making them computationally intensive.

CAP4D~\cite{taubner2024cap4d} is one exception that achieves real-time performance after an initial optimization phase. It directly encodes the input image into diffusion latents and conditions the model using 3D deformable model (3DMM) tracking data during avatar registration. Once optimized, the system can drive the avatar in real time.

In contrast, our model preserves the generative strengths of diffusion architectures while operating in a fully feed-forward manner. By refining pre-generated novel views in a single diffusion step, our method produces high-quality, 3D-aware portraits with minimal computational overhead.

\paragraph{Avatar generation.}
Another line of work uses pre-trained avatar generators to lift a reference image into 3D space and render it from a target camera viewpoint. Methods such as HideNeRF~\cite{li2023hidenerf}, Next3D~\cite{sun2023next3d}, OTAvatar~\cite{ma2023otavatar}, and GP-Avatar~\cite{chu2024gpavatar} follow this paradigm. These models employ feed-forward encoders to map the original image in the triplane space, thus avoiding costly optimization. For representing the expressions, HideNeRF and Next3D use deformation fields, while OTAvatar uses StyleGAN-based, decoupled expression latent control, and GP-Avatar uses point-based expression space. These methods rely on 3DMMs, defined with explicit human features like the ears, nose, and mouth, making the 3D avatar more realistic in shape when lifted from 2D inputs.

Although these models are computationally efficient and produce 3D-consistent outputs, they often compromise fidelity in preserving detailed identity and subtle expressions. In our work, we build on the speed and 3D awareness offered by avatar models while enhancing their identity and expression fidelity, resulting in high-quality, realistic novel views.

\paragraph{Image-to-image editing.}
Image-to-image (img2img) editing~\cite{rombach2021highresolution} has been widely adopted in diffusion-based models to modify an input image while preserving its overall structure. Unlike standard diffusion models that denoise from pure noise conditioned on a prompt, img2img starts from an existing image and gradually transforms it into a new version guided by textual or structural conditions. img2img-turbo~\cite{parmar2024img2imgturbo} improves efficiency by replacing the original Stable Diffusion model with Stable Diffusion Turbo and applying LoRA-based fine-tuning to reduce the number of trainable parameters. Difix3D~\cite{wu2025difix3d} extends this paradigm to novel view refinement, operating on NeRF-reconstructed scenes and using sparse reference views to improve consistency and quality.

In this work, we adopt a similar strategy to refine novel views generated by 3D-consistent synthesis models, enabling efficient, high-quality 3D-aware portrait generation from minimal input. Unlike Difix3D~\cite{wu2025difix3d}, which applies single-step refinement to general NeRF scenes, our framework is specifically designed for 3D-consistent portrait avatars, where preserving fine identity cues, hair details, and subtle expressions is critical.

\begin{figure*}[!htbp]
    \centering
    \includegraphics[width=0.90\textwidth,trim={0pt 0pt 0pt 0pt},clip]{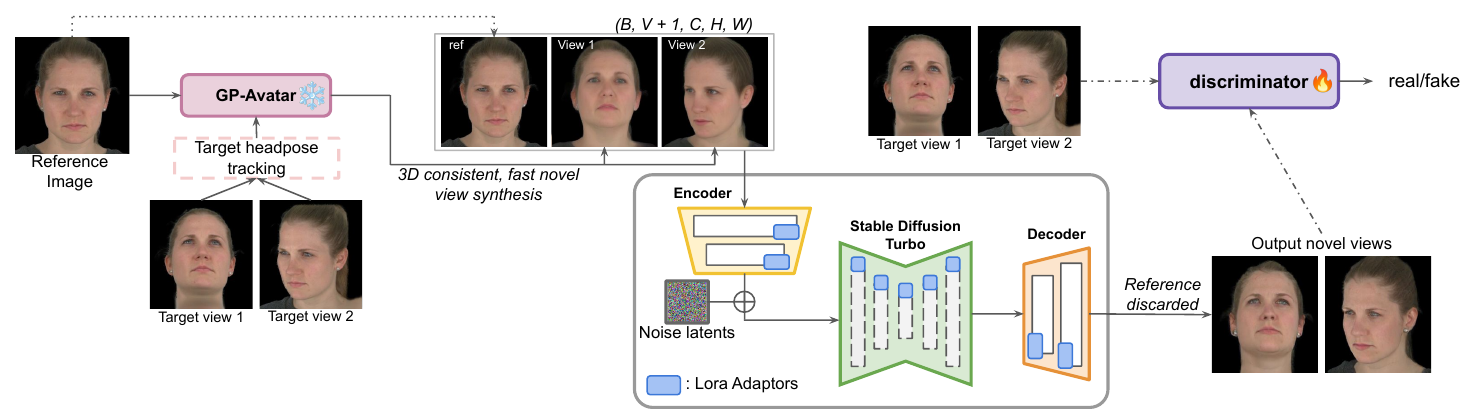}
    \caption{
 \textbf{TurboPortrait3D pipeline.} From a single reference image, GP-Avatar~\cite{chu2024gpavatar} generates 3D-consistent novel views from predicted target viewpoints. These, along with the reference image, are encoded into diffusion latents. Noise is added only to the novel views, and a Stable Diffusion Turbo U-Net with our attention-reshaping block refines them in a single step. The reference image is discarded, and the refined views are supervised with corresponding ground-truth images via a discriminator. Our design enables feed-forward, single-step, 3D-aware portrait refinement with high identity fidelity.
}
    \label{fig:architecture}
\end{figure*}

\begin{figure}[!htbp]
    \centering
    \includegraphics[width=0.25\textwidth,trim={0pt 0pt 0pt 0pt},clip]{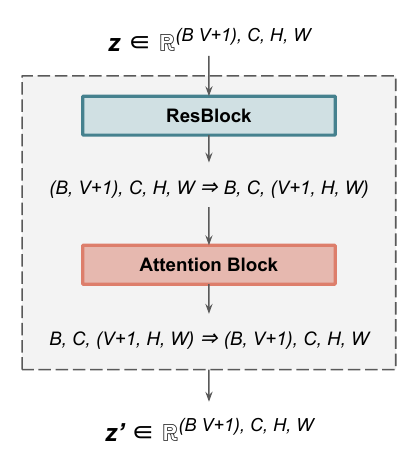}
    \caption{
  \textbf{Attention-reshaping block.} We reshape latents before and after the ResBlock and attention layers so identity and texture cues from the reference image propagate globally across all views, as introduced in Difix3D. This design achieves cross-view consistency without adding an extra U-Net for cross-attention, keeping the refinement lightweight and memory-efficient.
}
    \label{fig:diffusionblock}
\end{figure}

\section{Method}


We propose TurboPortrait3D, the first method specifically designed to perform single-step diffusion refinement for 3D-aware portrait avatars.

To generate novel views, we first synthesize a 3D avatar from a frontal reference image using GP-Avatar~\cite{chu2024gpavatar}. We then use the generated novel views, concatenated with the reference image, as input to our model to produce high-quality, 3D-aware results. Our model, illustrated in Figure~\ref{fig:architecture}, is inspired by Difix3D~\cite{wu2025difix3d} and relies on a single-step diffusion framework for image refinement.

\subsection{Preliminaries}

This subsection introduces the core components and prior works that form the basis of our method. We first present the avatar generation model used in our framework, GP-Avatar, which serves as the source of initial 3D-consistent novel view renders. We then review the relevant work on fast iterative refinement of noisy renders called Difix3D, which motivates our approach to efficient, high-quality novel view synthesis.

\paragraph{GP-Avatar.} GP-Avatar~\cite{chu2024gpavatar} generates a 3D avatar from one or more reference images by encoding the subject’s appearance into a tri-plane representation using a dedicated encoder. This architecture enables 3D-consistent avatar synthesis in a feed-forward manner, without requiring test-time optimization. GP-Avatar incorporates a point-based expression field, providing flexible control over facial expressions. For driving the avatar, we utilize target expressions and head poses extracted from ground-truth images, represented using FLAME~\cite{FLAME:SiggraphAsia2017} parameters. In our work, we focus specifically on manipulating head pose to enable novel view synthesis.

While GP-Avatar offers fast and 3D-consistent generation across various poses and expressions, it often struggles to preserve identity and fine-grained appearance details, particularly under large viewpoint changes. The generated outputs often exhibit visible blurring, particularly in the hair and side facial regions, which reduces overall photorealism. These limitations directly motivate our design of TurboPortrait3D, which introduces a multi-view-aware, attention-reshaping diffusion refinement stage that targets loss of identity cues and hair detail, capabilities not explored in prior single-step diffusion methods.

\paragraph{Difix3D.} Diffusion models generally require multiple denoising steps to progressively transform noisy latents into high-quality images. As a result, using fewer steps often produces blurry or underdefined outputs. Stable Diffusion Turbo~\cite{ADD} addresses this limitation by incorporating adversarial training with an auxiliary discriminator, enabling realistic image generation even with a reduced number of inference steps.

Difix3D~\cite{wu2025difix3d} builds upon this idea to refine NeRF-generated novel views of general 3D scenes in a single diffusion step, while preserving 3D consistency. It takes as input both novel views rendered from neural radiance fields, reconstructed from one or more reference images, and the reference images themselves. In contrast to other controllable diffusion models that utilize multiple U-Net modules with shared attention, Difix3D uses a lightweight architecture: it concatenates reference and novel views and applies view mixing across the diffusion blocks to promote self-supervised multi-view consistency. Refinement is performed using a single pre-trained U-Net from Stable Diffusion Turbo, resulting in fast and efficient inference. Unlike Difix3D, which operates on NeRF-reconstructed general scenes, our approach is specialized for portrait avatars generated by feed-forward parametric models, where cross-view identity consistency and recovery of high-frequency facial textures are critical because small deviations in facial geometry, hair flow, or expression can quickly break realism.

\subsection{TurboPortrait3D} \label{sec:methods_turboportrait3d}

Inspired by Difix3D, we adopt a similar strategy to enhance the quality of novel views generated by GP-Avatar. Our goal is to enable high-quality, feedforward portrait synthesis with improved identity preservation and 3D consistency.

First, we use GP-Avatar to generate $V$ 3D-consistent novel views $I^{v}_{GP}$ ($v \in \{1, \dots, V\}$) from a single frontal reference image $I_r$, using the viewpoints and head poses extracted from the target images $I^v$. These generated views are concatenated with the reference image to form the input to our model. We then map the input to our latent space $Z$ via a lightweight encoder.

\paragraph{Variable-noise training strategy.}
During training, we perturb each novel view latent $z_v \in Z$ with isotropic Gaussian noise $n \sim \mathcal{N}(0, \mathbf{1})$ scaled at varying levels $r \in \{0.0, 0.1, 0.2, \dots, 0.5\}$:
\begin{equation*}
z_v = (1 - r) \cdot z_v + r \cdot n.
\label{noise_latents}
\end{equation*}
This stochastic perturbation encourages the model to learn to recover novel views across a wide range of levels of input quality.
Importantly, we exclude the latent corresponding to the reference image from noise injection to preserve its original details. At inference time, we fix the noise level to $r = 0.1$ to ensure consistency across outputs. We report the ablation results for choosing this value in Table~\ref{tab:ablation_inference_noise}.

\paragraph{Memory-efficient refinement architecture.}
We pass the resulting noisy latent, with shape $(B, V+1, C, H, W)$, through a pre-trained U-Net model. As illustrated in Figure~\ref{fig:diffusionblock}, we first reshape the latent to $((B \cdot (V+1)), C, H, W)$ and process it using a ResBlock.
Then, we reshape the output to $(B, C, V+1, H, W)$ and rearrange it to $(B, C, (V+1 \cdot H \cdot W))$ for the attention block. This allows the model to propagate information from the reference image to the novel views, improving quality while preserving 3D consistency.
The reshaping ensures that identity and appearance cues from the reference image are globally accessible across all novel views, a property especially important in portrait synthesis where even small mismatches in facial geometry or texture are highly perceptible.

Finally, we pass the U-Net diffusion output, $\bar{I}^v$, to a discriminator, which provides adversarial supervision to further enhance photorealism.

Throughout training, we employ low-rank (LoRA)~\cite{hu2022lora} adapters to keep the model lightweight and efficient. The placement of the LoRA modules is shown in Figure~\ref{fig:architecture}. 

\section{Experiments}

\begin{figure*}[htbp!]
    \centering
    \includegraphics[width=0.8\textwidth,trim={0pt 0pt 0pt 0pt},clip]{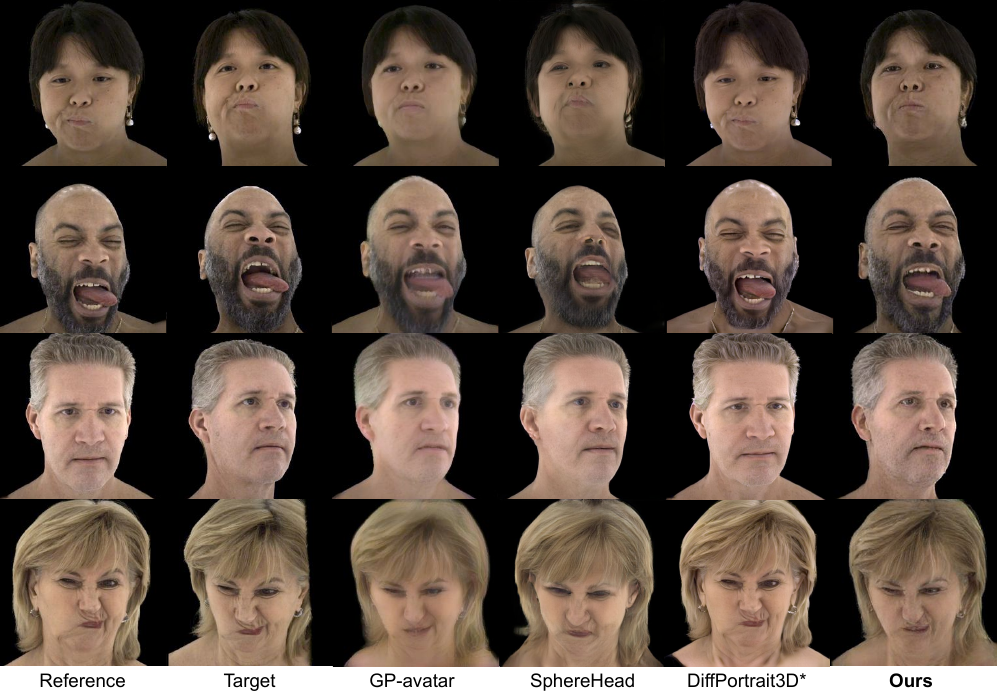}
    \caption{\textbf{Ava-256 validation results --} We report the qualitative results comparing state-of-the-art methods for 3D consistent, novel view portrait synthesis on Ava-256 validation data.}
    \label{fig:ava}
\end{figure*}

\begin{figure*}[htbp!]
    \centering
    \includegraphics[width=0.8\textwidth,trim={0pt 0pt 0pt 0pt},clip]{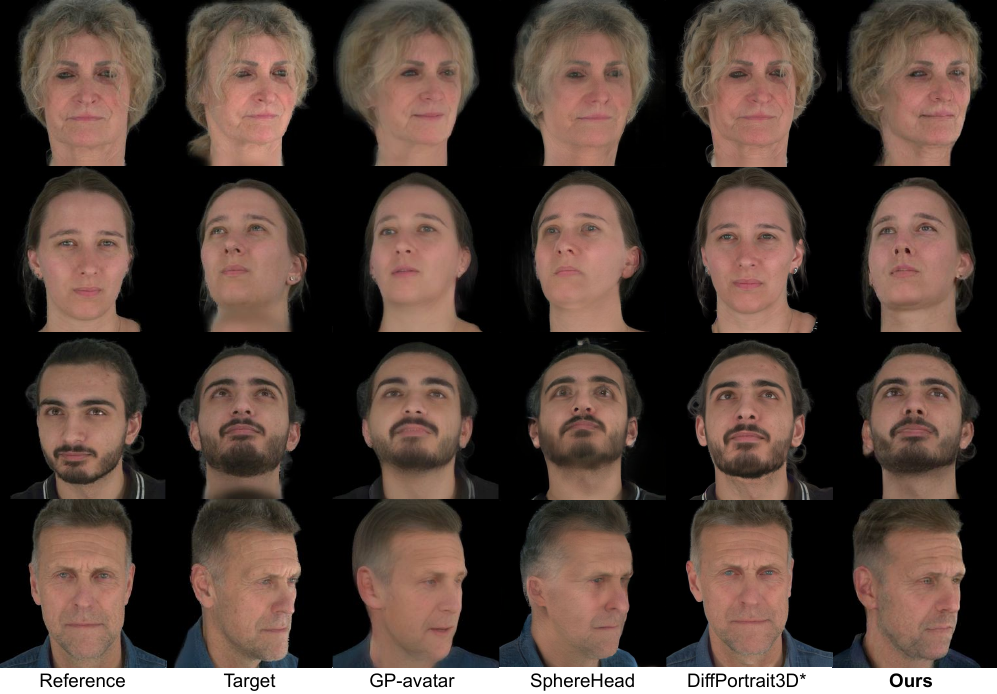}
    \caption{
    \textbf{NeRSemble validation results -- }We report the qualitative results comparing state-of-the-art methods for 3D consistent, novel view portrait synthesis on NeRSemble validation data.}
    \label{fig:nersemble}
\end{figure*}

\begin{table*}[!htbp]
    \centering
    \begin{tabular}{lrrrrrlrr}
        \toprule
        \textbf{Method}  & $\ell_2$ error $\downarrow$ & LPIPS $\downarrow$ & PSNR $\uparrow$ & SSIM $\uparrow$  &ID $\uparrow$ &FID$\downarrow$& \makecell{Registration \\ time (s) $\downarrow$} & \makecell{Generation \\ time (s) $\downarrow$} \\
        \midrule
     GP-Avatar &  \underline{0.019}& 0.252& 21.698& \underline{0.678}&\underline{0.899} &35.32&  3 & 0.07 \\
 SphereHead-PTI& 0.038& \underline{0.247}& \underline{21.741}& 0.662& 0.829 &\textbf{30.13}& 50 & 5.00\\
 DiffPortrait3D* (not aligned)& 0.052 & 0.281& 20.364& 0.633& 0.827 &30.38& * &28.00\\
 DiffPortrait3D* (aligned)& 0.047& 0.288& 20.751& 0.629& 0.799 &34.04& *& 28.00\\
 \midrule
 \textbf{Ours}& \textbf{0.013}& \textbf{0.213}& \textbf{23.143}&\textbf{ 0.698}&\textbf{0.949} &\underline{30.28}& 3 &0.12\\
 \bottomrule
    \end{tabular}
    \caption{
   \textbf{Quantitative analysis —} We present quantitative results comparing the performance of our model with several state-of-the-art methods on the combined Ava-256 and Nersemble datasets. The evaluation data is aligned using the same method employed during the training of our model. For DiffPortrait3D, this alignment corresponds to the ``not aligned" setting, as its model was trained with a different alignment strategy. For the ``aligned" DiffPortrait3D results, we apply the alignment method provided in its official codebase, which reflects the setup used during its training. We also report the registration times for GP-Avatar and SphereHead, as these models require this process at least once to generate views at various angles. DiffPortrait3D* does not disclose the registration method used; we used GP-Avatar which takes 3 seconds. For results by dataset, please refer to the the supplementary material.}~\label{tab:quantitative_analysis}
\end{table*}
\vspace{-2mm}

\textbf{Datasets.} We trained TurboPortrait3D using a combination of synthetic and real data to ensure robustness across diverse identities, expressions, and head poses. For synthetic data, we leveraged PanoHead, a 3D-aware portrait synthesis model, to generate images for 1,000 unique identities. For each identity, we synthesized three viewpoints: one frontal and two from different camera angles. PanoHead-generated outputs are inherently 3D-aware, which helps stabilize multi-view learning during pretraining. For real data, we used the NeRSemble and Ava-256 datasets, comprising 387 and 240 subjects, respectively. From each dataset, a subset of subjects (33 from NeRSemble and 16 from Ava-256) was held out for evaluation. All datasets were preprocessed with the same alignment, cropping, and normalization pipeline to ensure consistent input scale and framing across synthetic and real domains, enabling the model to generalize well to unseen datasets without additional fine-tuning.

For NeRSemble, we randomly sampled one to three frames per subject and selected seven novel camera viewpoints along with one frontal view as reference. For Ava-256, we followed a similar sampling strategy, selecting one to three frames per subject and up to 16 camera viewpoints in addition to the frontal reference view.

The training process was conducted in two stages: we first pre-trained the model using the synthetic dataset to establish generalizable features, and then fine-tuned it on the real data to improve performance on real-world identities and capture conditions.

\paragraph{Training details and efficiency.} We initialized the U-Net model with the pre-trained weights from Stable Diffusion Turbo. As described in Section~\ref{sec:methods_turboportrait3d}, we trained the model using LoRA to significantly reduce the number of trainable parameters and improve efficiency. During training, we used a fixed diffusion timestep of $t = 400$ for the U-Net. The model was trained on two NVIDIA GeForce RTX 4090 GPUs, each with 24~GB of memory. For efficiency reporting, we separately measure (1) registration time (avatar generation) and (2) refinement time (diffusion enhancement). With refinement taking less than X seconds per view, the method is practical for interactive use cases.

We set the number of novel views to $V = 2$, and each image was resized to a resolution of $256 \times 256$.

\subsection{Quantitative Results}

We compare our method against several state-of-the-art approaches for 3D-aware novel view portrait synthesis, including GP-Avatar~\cite{chu2024gpavatar}, SphereHead~\cite{li2024spherehead}, and DiffPortrait3D~\cite{gu2024diffportrait3d}. GP-Avatar and SphereHead are evaluated using their official pretrained models, while DiffPortrait3D* is our reimplementation without the unreleased 3D-aware noise module.

As the 3D-aware noise generation module, responsible for producing noisy, 3D-aware novel views, is not publicly available in the DiffPortrait3D codebase, we use GP-Avatar-generated novel views as input to ensure a fair and consistent comparison. Hence, we denote our reproduction of DiffPortrait3D without its unreleased 3D-aware noise module as DiffPortrait3D*. Although DiffPortrait3D* is reported to handle non-aligned inputs, we evaluate both aligned and non-aligned configurations in Table~\ref{tab:quantitative_analysis}.

Specifically, the ``not aligned" setting refers to images aligned using the same method employed in PanoHead~\cite{an2023panohead}, which was also used during the training of our model. In contrast, the ``aligned" setting uses the alignment protocol from EG3D~\cite{Chan2021}, which is consistent with DiffPortrait3D*'s training setup. We evaluate all models using a comprehensive set of metrics: L2 error, LPIPS, PSNR, SSIM, identity consistency~\cite{deng2019arcface}, FID, and both registration and inference time.

As expected, GP-Avatar achieves significantly faster registration times than SphereHead, while delivering comparable results across most quality metrics. This result justifies our choice of GP-Avatar-generated novel views as inputs to our model. Overall, our method consistently outperforms all baselines across the evaluated metrics, combining strong 3D-awareness with high perceptual fidelity even in single-step refinement. When compared against a re-implementation of DiffPortrait3D without its unreleased 3D-aware noise (DiffPortrait3D*), TurboPortrait3D delivers lower latency and higher perceptual quality under a shared input pipeline.

\subsection{Qualitative Results}

For qualitative analysis, we present results on the Ava-256 and NeRSemble evaluation datasets in Figures~\ref{fig:ava} and~\ref{fig:nersemble}. In both cases, we observe that GP-Avatar and SphereHead frequently fail to preserve fine appearance details, such as hair structure and facial wrinkles, resulting in noticeably blurrier outputs, particularly in regions where multi-view consistency is crucial for realism. These artifacts are particularly evident in challenging regions such as hair boundaries, fine wrinkles, and extreme yaw poses.

DiffPortrait3D* (aligned with our inference setup) shows poor alignment in both datasets; however, it generates high-quality images that preserve identity well, effectively capturing extreme expressions and unique appearance characteristics of the subject.

Our results demonstrate strong alignment, benefiting from the use of GP-Avatar outputs as input, which ensures 3D-awareness. This combination enables sharp facial contours, stable texture reproduction, and reduced ghosting even under large pose variations.

Notably, despite operating with only a single diffusion step, our method successfully captures the subject's detailed appearance attributes, combining identity fidelity with efficient inference.

\begin{figure*}[ht!]
    \centering
    \includegraphics[width=0.73\textwidth,trim={0pt 0pt 0pt 0pt},clip]{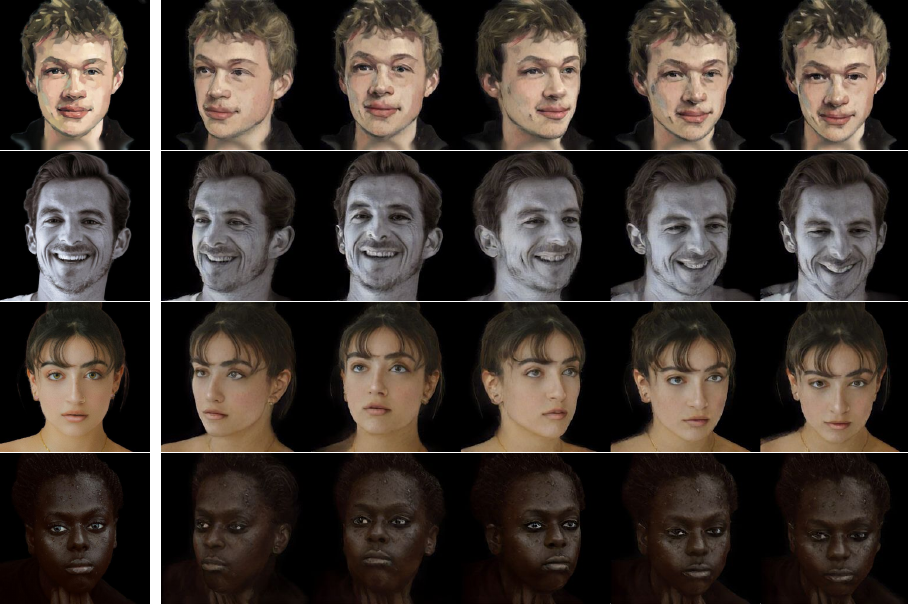}
    \caption{
\textbf{In-the-wild results —} We present additional qualitative results on in-the-wild images, including both real-world photographs and artistic renderings, to demonstrate the generalization ability of our model beyond curated datasets. The left-most image is the input image, and the images on the right are images generated from novel viewpoints.}
    \label{fig:itw}
\end{figure*}

We conclude by presenting qualitative results on in-the-wild images, including both real photographs and artistic renderings, as shown in Figure~\ref{fig:itw}. These examples demonstrate the robustness of our method to diverse, unconstrained inputs. Results can also be seen as videos in the Supplementary Material. 

\subsection{Ablation}

\begin{table}[h]
    \centering
    \small
    \begin{tabular}{lrrrr}
        \toprule
        \makecell{\textbf{Noise level}}& LPIPS $\downarrow$ & PSNR $\uparrow$ & SSIM $\uparrow$  &ID $\uparrow$\\
        \midrule
     0.0& 0.213& 23.114& 0.687&0.948\\
 0.1& 0.213& \textbf{23.143}& \textbf{0.698}&\textbf{0.949}\\
 0.2& \textbf{0.212}& 23.137& \textbf{0.698}& \textbf{0.949}\\
 0.3& 0.213& 23.116& 0.696& 0.949\\
 0.4& 0.213& 23.079& 0.695&0.947\\
 0.5& 0.216& 22.974& 0.692& 0.947\\
        \bottomrule
    \end{tabular}
    \caption{
    \textbf{Ablation on fixed noise levels during inference time --} We analyzed the performance of the model on fixed noise levels during inference.}~\label{tab:ablation_inference_noise}
\end{table}

\begin{figure}[t]
    \centering
    \includegraphics[width=0.45\textwidth,trim={0pt 0pt 0pt 0pt},clip]{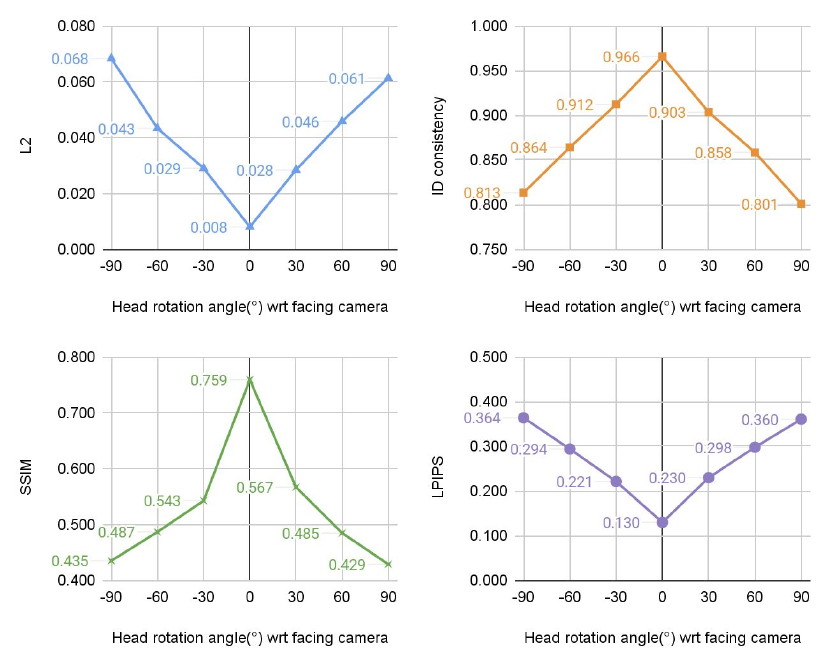}
    \caption{
    \textbf{Ablation on head rotation --} We report the performance on PanoHead images rotated around 180\textdegree. }
    \label{fig:ablation-angles}
\end{figure}

Since single-step refinement models can be sensitive to input perturbations, we evaluate model performance under different fixed latent noise levels during inference. Table~\ref{tab:ablation_inference_noise} reports results for LPIPS, PSNR, SSIM, and identity consistency across varying noise levels. We observe that the model remains robust, showing consistent performance across all metrics regardless of the noise level. We attribute this robustness to the stochastic latent perturbation strategy during training, which forces the model to recover from varying levels of degradation.

For our final results, we use a fixed noise level of 0.1, which yields the best overall performance.

We further ablate model performance across varying head poses by horizontally rotating the head over a 180\textdegree range. Specifically, we evaluate the model at seven angles: -90\textdegree, -60\textdegree, -30\textdegree, 0\textdegree, 30\textdegree, 60\textdegree, and 90\textdegree. Figure\ref{fig:ablation-angles} presents the results in terms of SSIM, LPIPS, identity consistency, and L2 error. As the head pose deviates further from the frontal view, we observe a consistent decline in performance for all metrics in both directions. 

While TurboPortrait3D achieves high identity preservation and realism, quality degrades at extreme yaw angles where GP-Avatar inputs lack sufficient side-view detail. Addressing these cases may require explicit hallucination or multi-view completion strategies.
\section{Discussion \& Limitations}

From our experimental results, we observe that TurboPortrait3D achieves strong performance while maintaining high time and memory efficiency. These advantages hold consistently across both controlled benchmark settings and in-the-wild scenarios, supporting the method's generalization ability. In contrast, DiffPortrait3D* exhibits notably lower performance than reported in the original paper~\cite{gu2024diffportrait3d}. We attribute this discrepancy to the absence of the 3D-consistent noise generation module in the publicly released codebase, an essential component used during its training.

Although TurboPortrait3D and DiffPortrait3D* share a similar architectural pipeline, both taking pre-generated 3D-consistent novel views as input to produce high-quality outputs, our method offers significant advantages in efficiency and deployability. TurboPortrait3D operates with a single diffusion step during inference, whereas DiffPortrait3D* requires 50 iterative denoising steps, greatly increasing computation time. In addition, TurboPortrait3D is lighter in memory usage, with a compact model design and fewer trainable parameters. It relies on a single UNet module and leverages LoRA for efficient fine-tuning. In contrast, DiffPortrait3D's larger memory footprint makes it infeasible to run on standard commercial GPUs, limiting its practical use in real-world scenarios.

Despite these advantages, our model has a few limitations. As shown in the ablation study in Figure~\ref{fig:ablation-angles}, performance degrades as the head pose deviates from the frontal view, particularly under extreme angles. This is partly due to the limited side-view detail in the GP-Avatar inputs, suggesting future improvements could focus on multi-view completion or explicit side-view hallucination. Additionally, our experiments were conducted with a limited number of novel view inputs ($V=2$) due to memory constraints. However, our framework is flexible and could support a larger number of input views to better capture the subject’s appearance in future extensions. Also, our DiffPortrait3D* results omit the method’s unreleased 3D-aware noise, which likely reduces its peak quality; our conclusions therefore emphasize latency and robustness under a shared, public codepath rather than claiming an absolute quality upper bound, in line with fair comparison principles.

In future work, we aim to extend TurboPortrait3D from single-frame refinement to video-based applications by incorporating temporal consistency constraints. One promising approach is to integrate temporal attention layers into the diffusion U-Net, allowing the model to exchange information between consecutive frames. This would help maintain stable identity features, hair flow, and expression dynamics over time, thereby reducing flicker and jitter in long sequences. Beyond temporal stability, we plan to explore fine-grained control mechanisms, such as parameterized facial rigs or learned expression embeddings, to enable interactive manipulation of expressions and poses. Coupled with our model’s existing 3D-awareness and identity preservation, these capabilities could make TurboPortrait3D suitable for real-time telepresence, VR avatars, and other interactive digital human systems, expanding its applicability well beyond static image synthesis.

{
    \small
    \bibliographystyle{ieeenat_fullname}
    \bibliography{main}

\begin{thebibliography}{21}
\providecommand{\natexlab}[1]{#1}
\providecommand{\url}[1]{\texttt{#1}}
\expandafter\ifx\csname urlstyle\endcsname\relax
  \providecommand{\doi}[1]{doi: #1}\else
  \providecommand{\doi}{doi: \begingroup \urlstyle{rm}\Url}\fi

\bibitem[An et~al.(2023)An, Xu, Shi, Song, Ogras, and Luo]{an2023panohead}
Sizhe An, Hongyi Xu, Yichun Shi, Guoxian Song, Umit Ogras, and Linjie Luo.
\newblock Panohead: Geometry-aware 3d full-head synthesis in 360$^{\circ}$, 2023.

\bibitem[Chan et~al.(2021)Chan, Lin, Chan, Nagano, Pan, Mello, Gallo, Guibas, Tremblay, Khamis, Karras, and Wetzstein]{Chan2021}
Eric~R. Chan, Connor~Z. Lin, Matthew~A. Chan, Koki Nagano, Boxiao Pan, Shalini~De Mello, Orazio Gallo, Leonidas Guibas, Jonathan Tremblay, Sameh Khamis, Tero Karras, and Gordon Wetzstein.
\newblock Efficient geometry-aware {3D} generative adversarial networks.
\newblock In \emph{arXiv}, 2021.

\bibitem[Chu et~al.(2024)Chu, Li, Zeng, Yang, Lin, Liu, and Harada]{chu2024gpavatar}
Xuangeng Chu, Yu Li, Ailing Zeng, Tianyu Yang, Lijian Lin, Yunfei Liu, and Tatsuya Harada.
\newblock {GPA}vatar: Generalizable and precise head avatar from image(s).
\newblock In \emph{The Twelfth International Conference on Learning Representations}, 2024.

\bibitem[{Content Authenticity Initiative}(2023)]{cai2023faces}
{Content Authenticity Initiative}.
\newblock How realistic are ai-generated faces?
\newblock \url{https://contentauthenticity.org/blog/how-realistic-are-ai-generated-faces}, 2023.
\newblock Accessed: 2025-08-09.

\bibitem[Deng et~al.(2019)Deng, Guo, Xue, and Zafeiriou]{deng2019arcface}
Jiankang Deng, Jia Guo, Niannan Xue, and Stefanos Zafeiriou.
\newblock Arcface: Additive angular margin loss for deep face recognition.
\newblock In \emph{Proceedings of the IEEE/CVF Conference on Computer Vision and Pattern Recognition}, pages 4690--4699, 2019.

\bibitem[Gu et~al.(2024)Gu, Xu, Xie, Song, Shi, Chang, Yang, and Luo]{gu2024diffportrait3d}
Yuming Gu, Hongyi Xu, You Xie, Guoxian Song, Yichun Shi, Di Chang, Jing Yang, and Lingjie Luo.
\newblock Diffportrait3d: Controllable diffusion for zero-shot portrait view synthesis.
\newblock In \emph{Proceedings of the IEEE Conference on Computer Vision and Pattern Recognition (CVPR)}, 2024.

\bibitem[Hu et~al.(2022)Hu, Shen, Wallis, Allen-Zhu, Li, Wang, Wang, and Chen]{hu2022lora}
Edward~J Hu, Yelong Shen, Phillip Wallis, Zeyuan Allen-Zhu, Yuanzhi Li, Shean Wang, Lu Wang, and Weizhu Chen.
\newblock Lo{RA}: Low-rank adaptation of large language models.
\newblock In \emph{International Conference on Learning Representations}, 2022.

\bibitem[Kim et~al.(2021)Kim, Chung, Park, Gu, Nam, Choe, Lee, and Choo]{kim2021k}
Taewoo Kim, Chaeyeon Chung, Sunghyun Park, Gyojung Gu, Keonmin Nam, Wonzo Choe, Jaesung Lee, and Jaegul Choo.
\newblock K-hairstyle: A large-scale korean hairstyle dataset for virtual hair editing and hairstyle classification.
\newblock In \emph{2021 IEEE International Conference on Image Processing (ICIP)}, pages 1299--1303. IEEE, 2021.

\bibitem[Li et~al.(2024)Li, Chen, Shi, Qiu, An, Chen, and Han]{li2024spherehead}
Heyuan Li, Ce Chen, Tianhao Shi, Yuda Qiu, Sizhe An, Guanying Chen, and Xiaoguang Han.
\newblock Spherehead: Stable 3d full-head synthesis with spherical tri-plane representation, 2024.

\bibitem[Li et~al.(2017)Li, Bolkart, Black, Li, and Romero]{FLAME:SiggraphAsia2017}
Tianye Li, Timo Bolkart, Michael.~J. Black, Hao Li, and Javier Romero.
\newblock Learning a model of facial shape and expression from {4D} scans.
\newblock \emph{ACM Transactions on Graphics, (Proc. SIGGRAPH Asia)}, 36\penalty0 (6):\penalty0 194:1--194:17, 2017.

\bibitem[Li et~al.(2023)Li, Zhang, Wang, Zhao, Wang, Chen, Zhang, Wang, Bo, and Li]{li2023hidenerf}
Weichuang Li, Longhao Zhang, Dong Wang, Bin Zhao, Zhigang Wang, Mulin Chen, Bang Zhang, Zhongjian Wang, Liefeng Bo, and Xuelong Li.
\newblock One-shot high-fidelity talking-head synthesis with deformable neural radiance field.
\newblock 2023.

\bibitem[Ma et~al.(2023)Ma, Zhu, Qi, Lei, and Zhang]{ma2023otavatar}
Zhiyuan Ma, Xiangyu Zhu, Guojun Qi, Zhen Lei, and Lei Zhang.
\newblock Otavatar: One-shot talking face avatar with controllable tri-plane rendering.
\newblock \emph{arXiv preprint arXiv:2303.14662}, 2023.

\bibitem[Paraperas~Papantoniou et~al.(2024)Paraperas~Papantoniou, Lattas, Moschoglou, Deng, Kainz, and Zafeiriou]{paraperas2024arc2face}
Foivos Paraperas~Papantoniou, Alexandros Lattas, Stylianos Moschoglou, Jiankang Deng, Bernhard Kainz, and Stefanos Zafeiriou.
\newblock Arc2face: A foundation model for id-consistent human faces.
\newblock In \emph{Proceedings of the European Conference on Computer Vision (ECCV)}, 2024.

\bibitem[Parmar et~al.(2024)Parmar, Park, Narasimhan, and Zhu]{parmar2024img2imgturbo}
Gaurav Parmar, Taesung Park, Srinivasa Narasimhan, and Jun-Yan Zhu.
\newblock One-step image translation with text-to-image models, 2024.

\bibitem[Roich et~al.(2021)Roich, Mokady, Bermano, and Cohen-Or]{roich2021pivotal}
Daniel Roich, Ron Mokady, Amit~H Bermano, and Daniel Cohen-Or.
\newblock Pivotal tuning for latent-based editing of real images.
\newblock \emph{ACM Trans. Graph.}, 2021.

\bibitem[Rombach et~al.(2021)Rombach, Blattmann, Lorenz, Esser, and Ommer]{rombach2021highresolution}
Robin Rombach, Andreas Blattmann, Dominik Lorenz, Patrick Esser, and Björn Ommer.
\newblock High-resolution image synthesis with latent diffusion models, 2021.

\bibitem[Sauer et~al.(2024)Sauer, Lorenz, Blattmann, and Rombach]{ADD}
Axel Sauer, Dominik Lorenz, Andreas Blattmann, and Robin Rombach.
\newblock Adversarial diffusion distillation.
\newblock In \emph{ECCV}, 2024.

\bibitem[Sun et~al.(2023)Sun, Wang, Wang, Li, Zhang, Zhang, and Liu]{sun2023next3d}
Jingxiang Sun, Xuan Wang, Lizhen Wang, Xiaoyu Li, Yong Zhang, Hongwen Zhang, and Yebin Liu.
\newblock Next3d: Generative neural texture rasterization for 3d-aware head avatars.
\newblock In \emph{CVPR}, 2023.

\bibitem[Taubner et~al.(2024)Taubner, Zhang, Tuli, and Lindell]{taubner2024cap4d}
Felix Taubner, Ruihang Zhang, Mathieu Tuli, and David~B. Lindell.
\newblock {CAP4D}: Creating animatable {4D} portrait avatars with morphable multi-view diffusion models.
\newblock 2024.

\bibitem[Wu et~al.(2025)Wu, Zhang, Turki, Ren, Gao, Shou, Fidler, Gojcic, and Ling]{wu2025difix3d}
Jay~Zhangjie Wu, Yuxuan Zhang, Haithem Turki, Xuanchi Ren, Jun Gao, Mike~Zheng Shou, Sanja Fidler, Zan Gojcic, and Huan Ling.
\newblock Difix3d+: Improving 3d reconstructions with single-step diffusion models.
\newblock \emph{arXiv preprint arXiv: 2503.01774}, 2025.

\bibitem[Zhang et~al.(2023)Zhang, Rao, and Agrawala]{zhang2023adding}
Lvmin Zhang, Anyi Rao, and Maneesh Agrawala.
\newblock Adding conditional control to text-to-image diffusion models, 2023.

\end{thebibliography}
}

\end{document}